%% file: arxiv.tex
\documentclass[letterpaper, 10 pt, conference]{ieeeconf}

\IEEEoverridecommandlockouts
\usepackage{times}
\usepackage{amsmath, amssymb}
\usepackage{graphicx}
\usepackage{booktabs}
\usepackage{url}
\usepackage{xcolor}
\usepackage{multirow}
\usepackage{cite}

\title{\LARGE \bf
EndoMD-SLAM: Endoscopic Gaussian Splatting SLAM under Optical Degradation with Memory and Static-Transient Decomposition
}

\author{Nuo Chen$^{*}$, Kangqi Ni$^{*}$, Lulin Liu, Joga Ivatury, Ying Ding,\\
Farshid Alambeigi, Tianlong Chen, Zhiwen Fan\\
[0.5em]
\small\textbf{Project Website:} \url{https://endomd-slam.github.io/}\thanks{$^{*}$These authors contributed equally.}
\thanks{Nuo Chen and Zhiwen Fan are with Texas A\&M University, College Station,
TX, USA. Kangqi Ni and Tianlong Chen are with the University of North Carolina
at Chapel Hill, Chapel Hill, NC, USA. Lulin Liu is with Texas A\&M University,
College Station, TX, USA, and the University of Minnesota, Minneapolis, MN, USA.
Joga Ivatury, Ying Ding, and Farshid Alambeigi are with the University of Texas
at Austin, Austin, TX, USA.}}

\makeatletter
\def\fnum@table{TABLE~\thetable}

\long\def\@makecaption#1#2{\vskip\abovecaptionskip
\footnotesize
\sbox\@tempboxa{#1:\hspace{0.5em}#2}\ifdim \wd\@tempboxa >\hsize
#1:\hspace{0.5em}#2\par
\else
\hbox to\hsize{\hfil\box\@tempboxa\hfil}\fi
\vskip\belowcaptionskip
}
\makeatother

\usepackage{caption}
\begin{document}

\IEEEaftertitletext{\vspace{-0.8\baselineskip}}

\maketitle

\thispagestyle{empty}
\pagestyle{empty}

\begin{abstract}
Dense 3D reconstruction is critical for clinical endoscopic navigation and documentation. While Gaussian Splatting SLAM systems show promise in this domain, they fundamentally rely on strict multi-view photometric consistency. In routine procedures, this assumption is severely violated by intermittent optical degradations like moving debris and water flushing. Standard systems erroneously fuse these camera-attached artifacts into the persistent 3D geometry, causing severe tracking drift and irreversible map corruption.
To address this limitation, we propose EndoMD-SLAM, a framework designed to maintain stability under optical degradation through specialized tracking and mapping mechanisms. On the tracking side, a memory-driven gating mechanism detects unreliable observations to suspend map updates and utilizes historical keyframes for drift-aware relocalization. On the mapping side, a self-supervised static-transient decomposition isolates visual contaminants into a dedicated transient field. This explicit separation prevents artifacts from structurally entangling with the persistent anatomical map.
We curate a degradation-focused benchmark from colonoscopy videos to systematically evaluate these failure modes. Extensive experiments show that while standard baselines fail under severe optical degradation, EndoMD-SLAM preserves geometric integrity, reducing absolute trajectory error by 91\% and improving rendering fidelity by 9.9 dB PSNR.
\end{abstract}

\input{sec/1_intro}

\input{sec/2_related_works}

\input{sec/3_method}

\input{sec/4_experiments}

\input{sec/5_conclusion}

\bibliographystyle{IEEEtran}
\bibliography{refs}

\end{document}

%% file: sec/1_intro.tex
\section{Introduction}

Endoscopy is a widely used, minimally invasive procedure that visualizes internal anatomy via an onboard camera for screening, diagnosis, and surgery~\cite{richter2024realtimeendo3d, ciuti2020frontiers}. Supporting these clinical applications requires reliable online tracking and high-quality 3D reconstruction, motivating increasing research efforts on endoscopic SLAM in recent years~\cite{azagra2023endomapper, wang2024endogslam, wu2025endoflow, liu2022sage, cao2025coarsetofinegsslam, kaleta2024prendo}. Traditional visual SLAM systems~\cite{grasa2013visual, mahmoud2017slam, wang2019visual, gomez2021sddefslam, teufel2024oneslam} reconstruct sparse point clouds while estimating camera poses. Although efficient, this sparse 3D information lacks the fine-grained surface detail required for clinical diagnosis and holistic anatomical review, posing the need for high-fidelity 3D map visualization during tracking. Recently, the field has moved toward dense, renderable mapping with Neural Radiance Fields (NeRFs)~\cite{mildenhall2022nerf}, which excel at novel view synthesis from the reconstructed 3D map. However, their implicit formulation introduces computational overhead that limits continuous online tracking in clinical applications. 3D Gaussian Splatting (3DGS)~\cite{kerbl2023gaussiansplatting} 
addresses this bottleneck by explicitly modeling scene geometry through efficient differentiable rasterization, enabling rapid dense reconstruction.

\begin{figure}[t]
\centering
\includegraphics[width=\columnwidth]{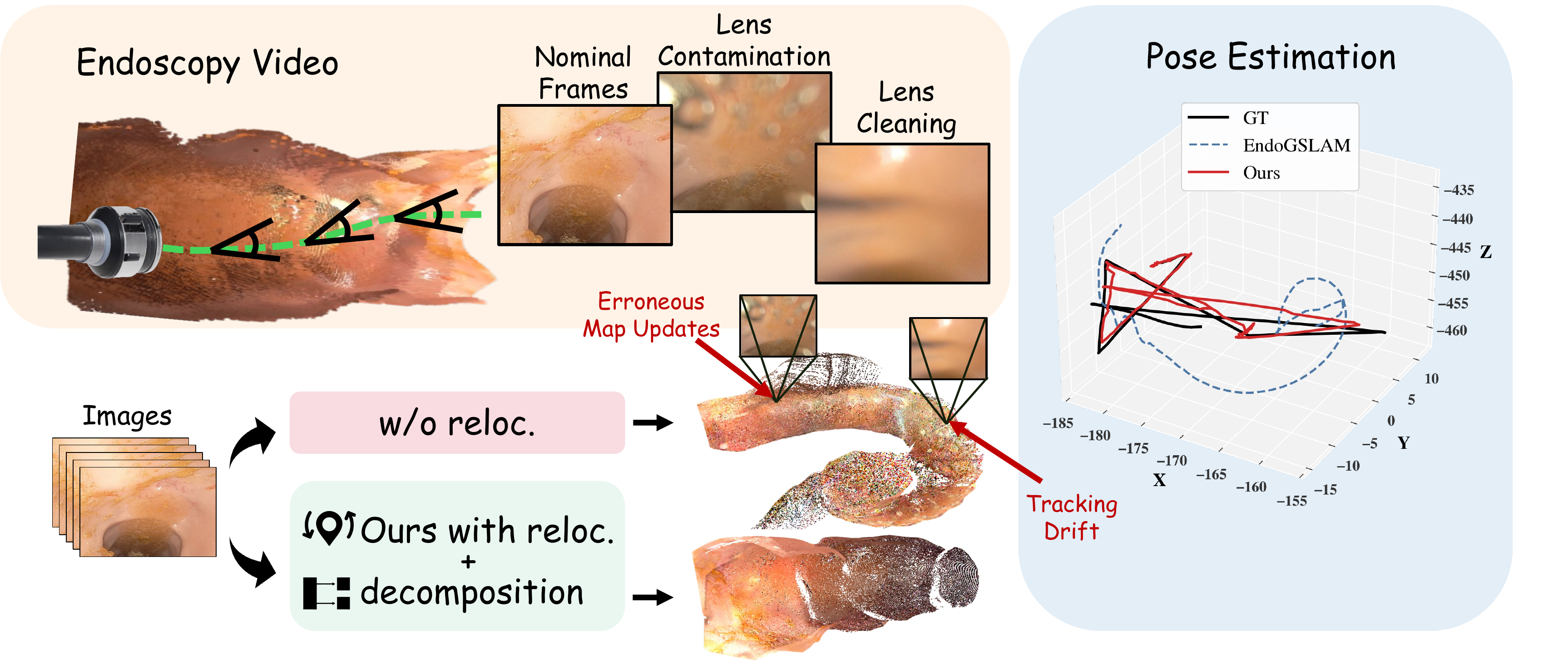}
\vspace{-6mm}
\caption{\textbf{SLAM under severe optical degradation.} Standard Gaussian Splatting SLAM systems fail under routine optical degradations, suffering severe tracking drift and erroneously mapping updates. With the proposed memory and decomposition mechanisms, EndoMD-SLAM isolates these degradations caused by contaminants and water flushing, maintaining robust online tracking and recovering a clean, high-fidelity 3D anatomical reconstruction.}
\label{fig:teaser}
\vspace{-6mm}
\end{figure}

Deploying 3DGS for SLAM assumes continuous, unobstructed visual observations, an assumption often challenged by the complex imaging conditions of clinical endoscopy (Fig.~\ref{fig:teaser}). In routine endoscopic practice, water droplets and biological debris adhere to the lens~\cite{yoshida2015risk}, creating optical degradations that compromise the camera's field of view. This necessitates lens cleaning via water flushing to restore visibility~\cite{yoshida2017novel} during the procedure. These clinically necessary events produce unstable visual evidence that directly violates the multi-view photometric consistency required by Gaussian Splatting SLAM. 
Intermittent degradations from flushing and transient artifacts from moving droplets disrupt frame-to-frame visual continuity. Existing endoscopic pipelines~\cite{azagra2023endomapper, wang2024endogslam, wu2025endoflow} often degrade under these conditions because they are primarily designed and evaluated on relatively clean endoscopic views, lacking dedicated mechanisms for optical degradation.

During tracking, water flushing causes sudden visibility loss that degrades observations and induces drift. This motivates mechanisms to pause the mapping process and recover from tracking failures; however, current systems~\cite{wang2024endogslam} still integrate corrupted poses, triggering erroneous map updates that permanently corrupt the global reconstruction. During mapping, transient droplets and debris act as camera-attached artifacts. When existing works directly fuse these transient observations, they erroneously reconstruct visual contaminants as persistent anatomical structure, which distorts rendering and further misguides tracking. Overcoming these coupled vulnerabilities requires explicitly triggering relocalization after degraded intervals and isolating transient artifacts from the persistent anatomy, alongside a memory mechanism to safely gate map updates.

To address this gap, we propose \textbf{\textit{EndoMD-SLAM}}, an endoscopic Gaussian splatting SLAM framework designed to maintain stability during colonoscopy procedures under optical degradation. Our approach combines a memory mechanism with a self-supervised static-transient decomposition. The temporal memory stores clean historical states, enabling drift-aware relocalization to recover pose information and gate Gaussian updates during degraded observations. Concurrently, the self-supervised static-transient decomposition separates a transient artifact field from a consistent static field. This prevents optical degradations from becoming structurally entangled with the static reconstruction (the persistent anatomy) during optimization. To validate our analysis and benchmark prior methods, we curate an optical degradation-focused benchmark from C3VDv2~\cite{golhar2025c3vdv2} to systematically evaluate tracking and mapping robustness.

Our main contributions are as follows:
\begin{itemize}
    \item We propose \textbf{\textit{EndoMD-SLAM}}, an endoscopic Gaussian splatting SLAM framework for colonoscopy 3D visualization under optical degradations in routine clinical practice. To prevent tracking drift and map corruption, we analyze the tracking and mapping vulnerabilities caused by these degradations and build a targeted benchmark suite.
          
    \item We introduce a temporal memory that maintains a bank of clean keyframes for tracking recovery and actively gates 3D Gaussian updates during low-confidence observations. By matching the current view to historical keyframes for relocalization, it recovers from tracking loss and prevents drift from corrupting the global map.
    
    \item We design a static-transient decomposition to separate persistent anatomy from transient artifacts, optimized in a self-supervised manner without requiring artifact masks or annotations, thereby preserving structural integrity and improving rendering fidelity.
    
    \item Extensive experiments systematically reveal common failure modes in tracking and mapping, demonstrating that our method outperforms prior approaches under optical degradation.
\end{itemize}

%% file: sec/2_related_works.tex
\section{Related Work}

\subsection{Endoscopic SLAM}
Endoscopic SLAM has evolved from adapting general-purpose visual SLAM pipelines to developing endoscopy-specific systems for tracking and dense reconstruction. Feature-based approaches build on classical foundations such as ORB-SLAM3~\cite{campos2021orb} for feature extraction, tracking, and relocalization, while endoscopy-specific systems such as CudaSIFT-SLAM~\cite{elvira2024cudasiftslam} improve full-procedure robustness in real colonoscopy through stronger features and multi-map recovery. Dense endoscopic mapping has advanced through learned depth estimation and neural rendering, including depth-and-motion pipelines~\cite{recasens2021endodepthmotion} and illumination-aware reconstruction such as LightNeuS~\cite{batlle2023lightneus}, and more recently through endoscopy-specific 3D Gaussian Splatting systems such as EndoGSLAM~\cite{wang2024endogslam} and EndoFlow-SLAM~\cite{wu2025endoflow} that enable real-time dense reconstruction and rendering. Complementary work has explored feature-based endoscopic VSLAM augmented with monocular depth for map densification~\cite{anadon2025densification}, while datasets such as EndoMapper~\cite{azagra2023endomapper} and C3VDv2~\cite{golhar2025c3vdv2} have strengthened evaluation for endoscopic mapping. Recent dense monocular SLAM systems such as MASt3R-SLAM~\cite{murai2025mast3rslam} show that learned 3D priors from geometric foundation model~\cite{wang2025vggt, lin2025depth, maggio2025vggt} can substantially strengthen monocular tracking and dense geometry recovery, but such systems primarily recover poses and dense point-based geometry rather than maintaining a renderable 3DGS scene model for novel-view synthesis.
In contrast, EndoMD-SLAM maintains drift-aware tracking and clean, renderable reconstructions by addressing the vulnerabilities of Gaussian Splatting SLAM under severe optical degradation.

\subsection{SLAM under Degraded Observations}
Robustness to visual degradation is a critical requirement in endoscopic SLAM, where lens flushing and transient contaminants violate standard photometric assumptions. Classical direct methods such as DSO~\cite{engel2018dso} highlight the sensitivity of photometric SLAM to illumination and appearance changes, motivating endoscopic approaches that treat degradation robustness as a first-class modeling issue; for example, NFL-BA~\cite{dunnbeltran2024nflba} replaces standard photometric bundle adjustment with a near-field-light-aware objective tailored to endoscopy’s moving light source. More broadly, dynamic-scene SLAM has long emphasized separating persistent structure from transient observations: Co-Fusion~\cite{runz2017cofusion} established this principle by jointly tracking and reconstructing multiple moving components, and recent dynamic 3DGS systems such as DG-SLAM~\cite{xu2024dgslam} and DGS-SLAM~\cite{kong2024dgsslam} extend it with motion masking, dynamic filtering, and robust or hybrid pose optimization. In medical and surgical imaging, non-static phenomena are further driven by deformable tissue and tools: NR-SLAM~\cite{rodriguez2024nr} models tissue deformation explicitly, while newer 3DGS-based surgical reconstruction methods such as Free-SurgS~\cite{guo2024freesurgs}, EndoGS~\cite{zhu2024endogs}, and Endo-4DGS~\cite{huang2024endo4dgs} extend Gaussian representations to deformable or time-varying surgical scenes. Dense SLAM systems such as NICE-SLAM~\cite{zhu2022niceslam}, Point-SLAM~\cite{sandstrom2023pointslam}, 
and Gaussian Splatting SLAM~\cite{matsuki2024gsslams} have also improved online neural and 3DGS mapping in general scenes, but are typically developed under substantially cleaner observation conditions than contamination-heavy endoscopy videos. 
Recent Gaussian rendering methods have increasingly explored decoupled scene representations~\cite{ding2026extrinsplat,wang2024desplat,lin2024hybridgs}, with DeSplat and HybridGS explicitly separating persistent scene structure from view-dependent or image-specific transients to improve robustness to distractors and contamination. However, these methods are formulated primarily for offline reconstruction or distractor-free novel-view synthesis from posed image sets rather than for online SLAM.
While robust rendering remains offline and online SLAM fails to recover tracking or prevent map corruption, EndoMD-SLAM bridges this gap via static-transient factorization, memory-driven gating, and relocalization.

%% file: sec/3_method.tex
\section{Method}
\label{sec_method}

\begin{figure*}[t]
\centering
\includegraphics[width=1\textwidth]{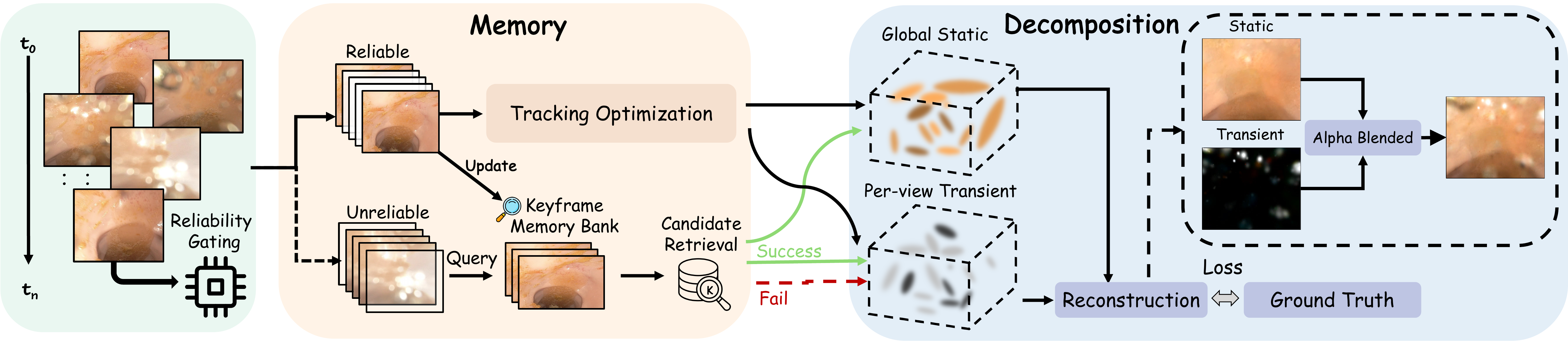}
\caption{\textbf{Overview of the EndoMD-SLAM architecture.} Input frames are first evaluated by a reliability gate. Non-degraded observations proceed to tracking optimization and update the global static map. Degraded frames activate the temporal memory, which queries a keyframe memory bank to perform drift-aware relocalization while safely suspending static map updates. Concurrently, a static-transient decomposition explicitly factorizes the scene. It absorbs camera-attached occluders into a per-view transient field, thereby protecting the persistent anatomy within the global static field. The static and transient renders are finally alpha-blended to reconstruct the degraded observation and compute the loss against the ground truth.}
\label{fig_overview}
\vspace{-5mm}
\end{figure*}

This section presents Endoscopic SLAM with Memory and Decomposition (EndoMD-SLAM), as illustrated in Fig.~\ref{fig_overview}. Section~\ref{sec_temporal} introduces the temporal memory for drift-aware relocalization and safe mapping. Section~\ref{sec_spatial} introduces the static-transient decomposition for separating a static anatomy field from a transient contamination field, which tightly couples with the temporal tracking process.

\subsection{Temporal memory for relocalization}
\label{sec_temporal}

Endoscopic videos contain intermittent degraded intervals caused by lens cleaning or sudden contamination. Under such degradations, pose estimates become unreliable. This is especially harmful for online Gaussian splatting SLAM because pose estimation is tightly coupled with continuous map updates. EndoMD-SLAM breaks this error accumulation loop by maintaining a temporal memory of unobstructed states to drive transient-aware tracking, perform relocalization, and safely gate map updates.

\paragraph{Transient-aware tracking}
To establish robust cross-frame correspondences in textureless endoscopic environments, 
we leverage correspondence priors from foundation geometric model~\cite{leroy2024mast3r} for matching and localization. In degraded endoscopic observations with severe appearance collapse and weak texture, conventional feature matching becomes unreliable, making foundation-model features necessary to maintain usable correspondences.
Given a reference keyframe $k$, the foundation model effectively lifts matched pixels to 3D pairs $\{(\mathbf{A}_i,\mathbf{B}_i)\}$ and assigns base confidence weights $w_i$. While this foundation-level matching provides exceptional robustness, it operates under a general static-scene assumption. Consequently, it accurately matches and assigns high confidence to moving lens contaminants (e.g., water droplets), inadvertently treating these camera-attached artifacts as valid 3D anatomical geometry.

To prevent these transient features from skewing the pose estimation, we introduce a transient-aware backend. Rather than utilizing the foundation model's outputs directly, we explicitly modulate the base weights $w_i$ using the transient opacity mask $\hat{\alpha}^{\mathrm{t}}_{t-1}$ rendered from our transient field. This yields a degradation-aware, re-weighted correspondence:
\begin{equation}
w_i' = w_i \cdot m(u_i), \qquad
m(u_i)=\max\!\big(1-\hat{\alpha}^{\mathrm{t}}_{t-1}(u_i),\,0\big)^{\gamma}.
\label{eq:reliability_weighting}
\end{equation}
If the static-transient decomposition has not yet initialized $\hat{\alpha}^{\mathrm{t}}_{t-1}$, we default to $m(u_i)=1$. We then estimate the relative camera pose by solving a depth-metric alignment:
\begin{equation}
T_{tk}^{\star}=\arg\min_{T\in \mathrm{SE}(3)} \sum_i w_i' \left\lVert \mathbf{B}_i - T\,\mathbf{A}_i \right\rVert_2^2 .
\label{eq:tracking_objective}
\end{equation}
The current pose is then updated as $T_t \leftarrow T_k\,T_{tk}^{\star}$. By solving Eq.~\eqref{eq:tracking_objective} in closed form via weighted Procrustes, we leverage the raw power of the foundation model's feature matching, but strictly filter the inputs through our transient mask. This guarantees a deterministic and stable pose update that is physically consistent with the static anatomy. This step yields the pose $T_t$ alongside our memory-triggering statistics: the valid correspondence fraction $\rho_t$ and the depth-valid pair count $N_{\mathrm{valid}}$.

\paragraph{Reliability gate}
Using the correspondence statistics derived from the tracking optimization, we evaluate the geometric consistency of the current observation. We define a binary reliability gate as:
\begin{equation}
g_t=\mathbb{I}\big[\rho_t\ge \tau_\rho \ \wedge\ N_{\mathrm{valid}}\ge \tau_n\big].
\label{eq:reliability_gate}
\end{equation}
EndoMD-SLAM uses $g_t$ to trigger memory-based recovery and to control whether static map updates are permitted.

\paragraph{Memory-based relocalization}
We maintain a temporal memory bank of reliable keyframes together with a retrieval database for instantaneous pose recovery. When the tracking estimate is deemed unreliable ($g_t=0$), the system must re-establish its pose without accumulating drift. Standard heuristic feature matching frequently fails to bridge the severe visual gaps caused by lens flushing. Therefore, to query our retrieval database, we leverage the robust dense descriptors extracted by the foundation model~\cite{leroy2024mast3r}. This allows us to reliably identify a localized candidate set of historical keyframes $\mathcal{C}_t$. 

For each candidate $k\in\mathcal{C}_t$, we re-run our transient-aware depth-metric solver to compute a pose hypothesis $T_t^{(k)}$ and evaluate its geometric match fraction $\rho_t^{(k)}$. We select the hypothesis that yields the highest consensus:
\begin{equation}
k^\star = \arg\max_{k\in\mathcal{C}_t}\ \rho_t^{(k)},\qquad T_t \leftarrow T_t^{(k^\star)} .
\label{eq:reloc_select_matchfrac}
\end{equation}
We accept the recovered pose only if it satisfies the strict reliability criteria established during tracking ($g_t=1$). Upon successful recovery, the frame is promoted to a new keyframe and inserted into the retrieval database. This ensures the temporal memory is continuously refreshed with validated, contamination-free states, enabling the system to survive intermittent blinding without relying on computationally prohibitive global map optimizations.

\paragraph{Safe mapping}
A key role of temporal memory is to prevent irreversible map corruption. Let $\Theta_{\mathrm{s}}$ denote the parameters of the static map. Using the reliability gate $g_t$, we suspend static updates on unreliable frames and only allow static integration when the pose is reliable:
\begin{equation}
\Theta_{\mathrm{s}}^{t+1} =
\begin{cases}
\Theta_{\mathrm{s}}^{t} - \eta \nabla_{\Theta_{\mathrm{s}}}\mathcal{L}_{\mathrm{map}}\!\left(\Theta_{\mathrm{s}}^{t}, \Theta_{\mathrm{t}}^{t};\, T_t, I_t, D_t\right), & g_t=1 \\
\Theta_{\mathrm{s}}^{t}, & g_t=0
\end{cases}
\label{eq:safe_static_gate}
\end{equation}
This safe mapping ensures that frames with unreliable poses do not write erroneous geometry into the static map. When $g_t=0$, we further allow transient-only fitting to absorb short-lived degradations while keeping the static map fixed.

\subsection{Static-transient decomposition}
\label{sec_spatial}

Lens droplets and debris introduce spatially localized transient distractors. Fusing them into a single map biases geometry and degrades tracking. EndoMD-SLAM addresses this by employing a static-transient decomposition that factorizes persistent anatomy from transient contamination, explicitly separating these artifacts into a transient field to prevent them from being fused into the static reconstruction.

\paragraph{Static-transient representation and compositing}
We represent the scene with two Gaussian fields. The static field models persistent anatomy, while the transient field absorbs lens artifacts. For a pose estimate at time $t$, we render the static RGB $\hat I_t^{\mathrm{s}}$ and the transient RGB $\hat I_t^{\mathrm{t}}$, alongside their respective silhouette alphas $\hat\alpha_t^{\mathrm{s}}$ and $\hat\alpha_t^{\mathrm{t}}$. The final RGB prediction is composed via an occlusion-style overlay:
\begin{equation}
\hat I_t
=
\hat I_t^{\mathrm{t}}
+
\bigl(1-\hat\alpha_t^{\mathrm{t}}\bigr)\odot \hat I_t^{\mathrm{s}}.
\label{eq:compose}
\end{equation}

\paragraph{Separation objective}
During mapping, we optimize the fields using a reconstruction loss $\mathcal{L}_{\mathrm{rgb}}$ and a depth loss $\mathcal{L}_{\mathrm{d}}$ that supervises only the static geometry:
\begin{equation}
\mathcal{L}_{\mathrm{base}}
=
\lambda_{\mathrm{rgb}}\mathcal{L}_{\mathrm{rgb}}(\hat I_t,I_t)
+
\lambda_{\mathrm{d}}\mathcal{L}_{\mathrm{d}}(\hat D_t^{\mathrm{s}},D_t).
\label{eq:base_loss}
\end{equation}
To encourage clean separation, we regularize transient occupancy and static coverage:
\begin{equation}
\mathcal{L}
=
\mathcal{L}_{\mathrm{base}}
+
w_{\mathrm{tr}}\,\mathbb{E}_u\!\left[\hat\alpha_t^{\mathrm{t}}(u)\right]
+
w_{\mathrm{bg}}\,\mathbb{E}_u\!\left[1-\hat\alpha_t^{\mathrm{s}}(u)\right].
\label{eq:sep_obj}
\end{equation}

\paragraph{Transient field initialization}
Each frame $t$ is assigned a fixed-capacity transient Gaussian set of $N_{\mathrm{t}}$ transient Gaussians. Upon first observation, we initialize the transient means by backprojecting sampled valid depth pixels to 3D:
\begin{equation}
\mu^{\mathrm{t}}_{t,i}
=
T_t^{-1}\!\left(D_t(u_i)\,K^{-1}\bar{\mathbf{u}}_i\right),
\qquad i=1,\ldots,N_{\mathrm{t}}.
\label{eq:transient_init}
\end{equation}
We initialize colors from the sampled RGB values with a small initial opacity. This anchors the transient slot to the current observation, allowing it to capture local artifacts independently of the global static field.

\paragraph{Transient field maintenance}
To keep the transient field adaptive, we periodically recycle low-contribution transient Gaussians in the transient field. We define a per-Gaussian activity score $a_{t,i}$ and mark inactive elements:
\begin{equation}
\mathcal{I}_t=\left\{i \,\middle|\, a_{t,i}<\tau_{\mathrm{act}}\right\}.
\label{eq:transient_recycle_set}
\end{equation}
Elements in $\mathcal{I}_t$ are replaced with duplicating informative elements or re-sampling. By explicitly managing the lifecycle of these slots, the transient field remains adaptive to evolving contaminants without accumulating stale artifacts.

%% file: sec/4_experiments.tex
\section{Experiments}
\label{sec_experiments}

We evaluate EndoMD-SLAM under severe optical degradation. We report tracking accuracy, reconstruction quality, and runtime, and provide qualitative analyses of static-transient decomposition and drift-aware relocalization.

\begin{table*}[!t]
\centering
\small
\setlength{\tabcolsep}{4pt}
\begin{tabular}{lccccc}
\toprule
Methods & PSNR $\uparrow$ & SSIM $\uparrow$ & LPIPS $\downarrow$ & RMSE(mm) $\downarrow$ & ATE (mm) $\downarrow$ \\
\midrule

MonoGS~\cite{matsuki2024gsslams} & 6.39 & 0.298 & 0.681 & 35.92 & 15.88 \\
EndoGSLAM~\cite{wang2024endogslam} & 8.16 & 0.335 & 0.628 & 26.21 & 35.91 \\
NICE-SLAM~\cite{zhu2022niceslam} & 14.20 & 0.406 & 0.725 & 20.01 & 13.57 \\

\midrule
EndoMD-SLAM & \textbf{18.06} & \textbf{0.606} & \textbf{0.416} & \textbf{7.22} & \textbf{3.02} \\
\bottomrule
\end{tabular}
\caption{
\textbf{Main results on degradation-focused benchmark.}
EndoMD-SLAM establishes a new state-of-the-art across all metrics under severe optical degradation. Best results are highlighted in \textbf{bold}.}
\label{tab:main_results}
\vspace{-1mm}
\end{table*}

\begin{table*}[!t]

\centering
\small
\setlength{\tabcolsep}{5pt}
\renewcommand{\arraystretch}{1.05}
\begin{tabular}{ccc|ccccc}
\toprule
Baseline & Memory & Decomposition & PSNR $\uparrow$ & SSIM $\uparrow$ & LPIPS $\downarrow$ & RMSE(mm) $\downarrow$ & ATE (mm) $\downarrow$ \\
\midrule
\checkmark &         &         & 8.16 & 0.335 & 0.628 & 26.21 & 35.91 \\
\checkmark & &\checkmark &          7.79 & 0.226 & 0.671 & 28.58 & 36.23 \\
\checkmark & \checkmark &         & 16.21 & 0.571 & 0.421 & \textbf{7.11}  & 3.04  \\
\checkmark & \checkmark & \checkmark& \textbf{18.06} & \textbf{0.606} & \textbf{0.416} & 7.22 & \textbf{3.02} \\
\bottomrule
\end{tabular}
\caption{\textbf{Quantitative ablation of our proposed components. }Both the temporal memory and spatial decomposition modules are critical for maximizing tracking robustness and rendering fidelity. Best results are highlighted in \textbf{bold}.}
\label{tab:ablations}

\end{table*}

\begin{figure*}[!t]
\centering
\includegraphics[width=0.99\textwidth]{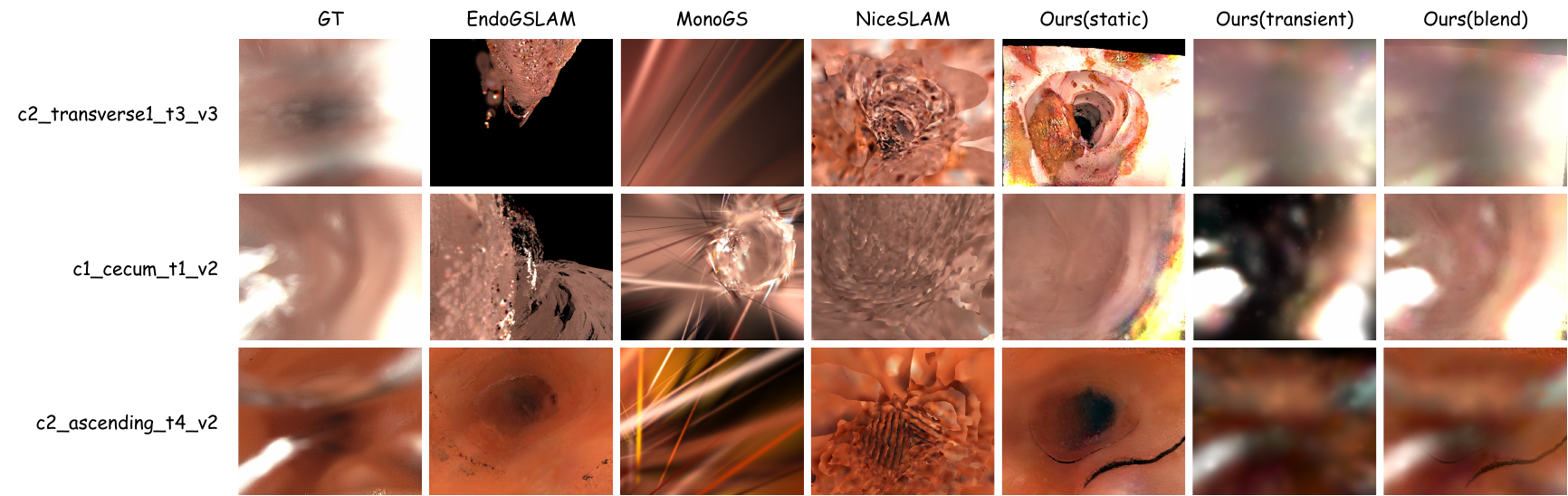}
\caption{\textbf{Qualitative comparison of novel view synthesis and field decomposition under severe optical degradation.} Standard baselines (EndoGSLAM, MonoGS, NICE-SLAM) exhibit prominent rendering artifacts and geometric distortions under degraded visual evidence. EndoMD-SLAM explicitly factorizes the scene via a static-transient decomposition. The transient field (Ours-transient) absorbs optical degradations, yielding a clean anatomical map (Ours-static), while the blended output (Ours-blend) reconstructs the degraded observation.}
\label{fig_qual_decomp}
\vspace{-5mm}

\end{figure*}

\subsection{Experimental setup}
\label{sec_exp_setup}

\paragraph{Degradation focused benchmark}
To evaluate tracking and reconstruction performance under severe lens obstruction, we constructed a targeted benchmark using the C3VDv2 dataset~\cite{golhar2025c3vdv2}. By filtering the dataset's metadata for sequences explicitly tagged with ``water on lens'' and ``debris on lens'', we identified ten highly degraded sequences. Following previous works~\cite{wang2024endogslam, wu2025endoflow}, all evaluations on these sequences were conducted at a resolution of $675 \times 540$.

\paragraph{Baselines} We compare our method against three representative state-of-the-art dense SLAM systems to ensure a comprehensive evaluation. These include NICE-SLAM~\cite{zhu2022niceslam} as a foundation NeRF-based approach, MonoGS~\cite{matsuki2024gsslams} representing general-purpose 3DGS SLAM, and EndoGSLAM~\cite{wang2024endogslam} as the leading endoscopy-specific 3DGS framework. To ensure a fair comparison, all baselines are evaluated using their official implementations under identical configurations on our curated benchmark.

\paragraph{Evaluation metrics} Following the standard evaluation protocols of recent endoscopic SLAM systems~\cite{wang2024endogslam, wu2025endoflow}, we evaluate tracking accuracy using Absolute Trajectory Error (ATE), geometric reconstruction using depth Root Mean Squared Error (RMSE)~\cite{sturm2012benchmark}, and novel view synthesis quality using PSNR, SSIM~\cite{wang2004image}, and LPIPS~\cite{zhang2018unreasonable}.

\paragraph{Implementation details}

We implement EndoMD-SLAM in PyTorch with CUDA and run all experiments on an NVIDIA RTX A6000 server. Following the standard experimental setting of~\cite{wu2025endoflow, zhu2024endogs}, our system utilizes RGB-D sequences as input. For the temporal memory, we retrieve up to $|\mathcal{C}_t| = 10$ candidates during relocalization. The tracking reliability gate enforces a minimum match fraction $\tau_{\rho} = 0.15$ and a minimum valid metric point count $\tau_n = 500$. For a fair rendering comparison with EndoGSLAM~\cite{wang2024endogslam}, we disable Spherical Harmonics optimization and run exactly 25 mapping iterations per frame to update the static and transient fields. We allocate $N_t = 1000$ transient Gaussians per frame and set the transient-aware tracking weight modulation exponent to $\gamma = 2.0$. Loss weights are $\lambda_{\mathrm{rgb}} = 1.0$, $\lambda_{\mathrm{d}} = 1.0$, $w_{\mathrm{tr}} = 0.01$, and $w_{\mathrm{bg}} = 0.01$.

\subsection{Main Results}
Tab.~\ref{tab:main_results} presents the quantitative comparison of all methods on our degradation-focused benchmark. Under severe lens obstruction, existing state-of-the-art systems experience significant tracking drift. Specifically, the general-purpose 3DGS SLAM framework, MonoGS~\cite{matsuki2024gsslams}, and the endoscopy-specific system, EndoGSLAM~\cite{wang2024endogslam}, yield Absolute Trajectory Errors (ATE) of 15.88~mm and 35.91~mm, respectively. These systems intrinsically assume continuous visual reliability; when sudden optical degradations interrupt this continuity, corrupted poses are inadvertently integrated into the trajectory. Similarly, the NeRF-based baseline, NICE-SLAM~\cite{zhu2022niceslam}, struggles to maintain accurate localization under these dynamic disturbances (13.57~mm ATE). By contrast, EndoMD-SLAM achieves a robust ATE of 3.02~mm. This stability is fundamentally enabled by our temporal memory, which acts as a safeguard to detect degraded frames, suspend map updates, and utilize stored reliable states to perform relocalization and recover the trajectory.

Furthermore, EndoMD-SLAM demonstrates superior performance in scene reconstruction and novel view synthesis. Our method achieves a PSNR of 18.06~dB and limits the depth RMSE to 7.22~mm. In this highly degraded regime, existing frameworks struggle considerably; EndoGSLAM yields 8.16~dB PSNR and 26.21~mm RMSE, while MonoGS drops to 6.39~dB PSNR. This performance gap occurs because baselines lack a mechanism to isolate camera-attached contaminants, frequently fusing transient artifacts such as water droplets and debris directly into the persistent 3D geometry. By employing a static-transient decomposition to explicitly factorize the scene into a static anatomy field and a transient artifact field, EndoMD-SLAM prevents these contaminants from corrupting the global map. This explicit decoupling results in structurally accurate and visually clean dense reconstructions, effectively maintaining the integrity of the anatomical mapping despite severe clinical degradations.

\subsection{Ablation Study}
To evaluate the individual contributions of our core components, we conduct an ablation study on the degradation-focused benchmark, calculating all metrics using the exact same sequences and evaluation protocols as the main experiments, with quantitative results detailed in Tab.~\ref{tab:ablations}.

\paragraph{Effectiveness of Temporal Memory} The temporal memory is designed to identify degraded observations, suspend map updates, and perform drift-aware relocalization. When this mechanism is disabled, the system cannot recover from tracking drifting during severe lens contamination, causing the Absolute Trajectory Error (ATE) to surge to 36.23~mm. Because the system forces these corrupted poses into the mapping pipeline, the global 3D reconstruction is severely compromised, resulting in a degraded PSNR of 7.79~dB. By introducing the temporal memory, EndoMD-SLAM successfully maintains trajectory stability (ATE 3.02~mm) and prevents erroneous geometry from corrupting the static map.

\paragraph{Effectiveness of Decomposition} The static-transient decomposition explicitly factorizes the scene into a persistent static anatomy field and a transient contamination field. When we ablate this module, the system still maintains a relatively robust trajectory (yielding an ATE of 3.04~mm) due to the active temporal memory mechanism. However, without static-transient decomposition, camera-attached contaminants such as water droplets and debris are inadvertently fused into the static 3D geometry. This geometric entanglement leads to a noticeable drop in rendering fidelity, reducing the PSNR from 18.06~dB to 16.21~dB. Integrating this decomposition effectively absorbs these local transient distractions, preserving the structural integrity and visual quality of the anatomical map.

\begin{figure}[!t]
\centering
\includegraphics[width=0.99\linewidth]{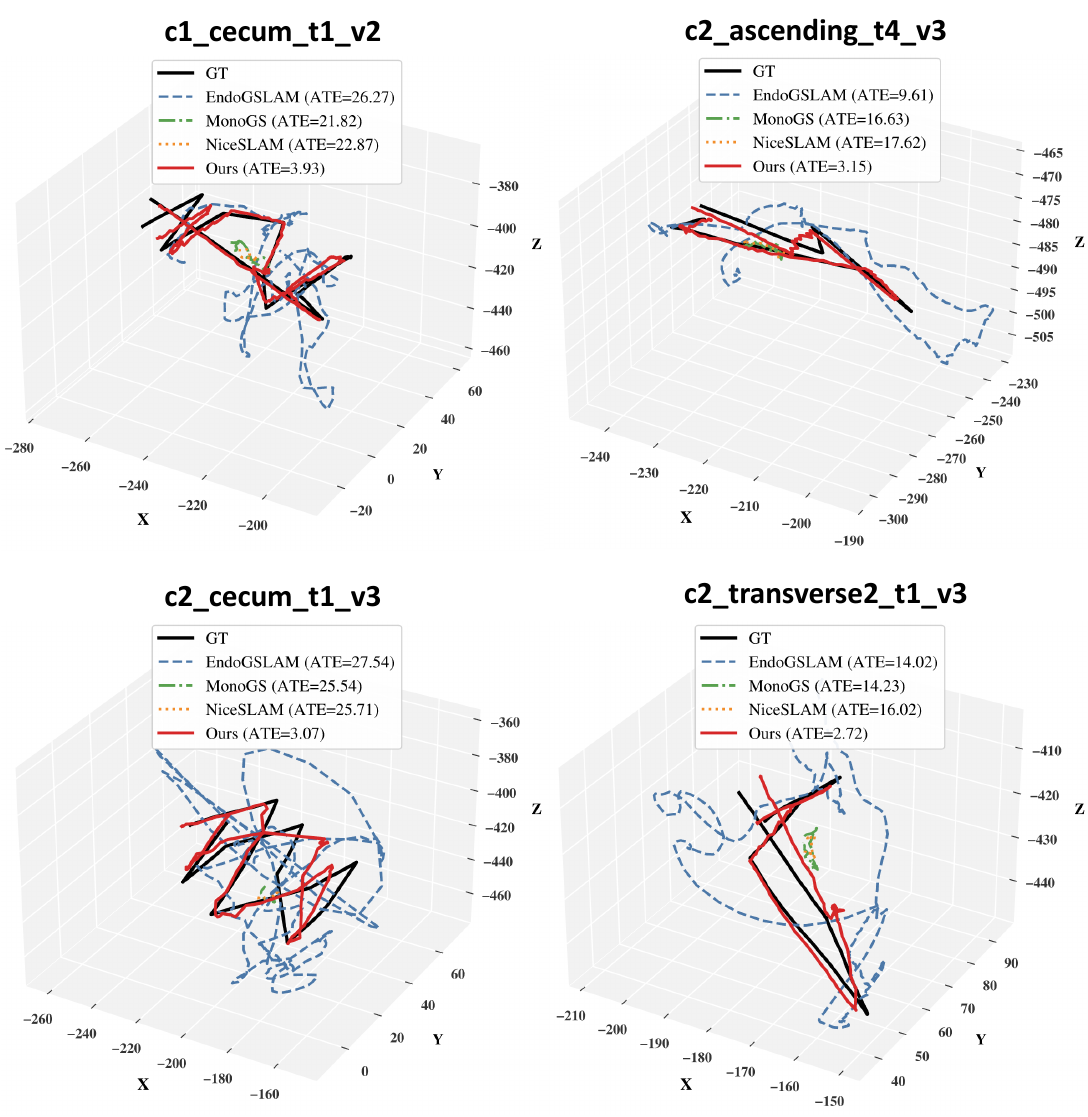}
\caption{\textbf{Qualitative comparison of estimated camera trajectories across diverse colonoscopy sequences.} Under severe lens contamination, the baseline method (blue dashed lines) experiences catastrophic tracking drift and frequent localization failures due to the sudden loss of visual continuity. In contrast, EndoMD-SLAM (red solid lines) maintains robust and accurate pose estimation that tightly aligns with the Ground Truth (black solid lines). By leveraging our temporal memory to gate map updates and perform failure-aware relocalization, our system successfully recovers from highly degraded intervals and prevents long-term trajectory divergence.}
\label{fig_qual_recovery}
\vspace{-5mm}
\end{figure}

\subsection{Qualitative Results}
\label{sec_qualitative}

We evaluate novel view synthesis and static-transient field decomposition, as shown in Fig.~\ref{fig_qual_decomp}. Standard mapping frameworks operate under the strict assumption of continuous photometric consistency. Consequently, when presented with localized lens contamination, MonoGS and EndoGSLAM attempt to reconstruct 2D water droplets and glare as physical 3D structures. This results in catastrophic rendering artifacts characterized by massive black voids and streaking artifacts that obscure the underlying anatomy. Similarly, the implicit representation of NICE-SLAM collapses under degraded multi-view constraints. EndoMD-SLAM bypasses this failure mode via explicit scene factorization. As shown in the rightmost columns of Fig.~\ref{fig_qual_decomp}, our static-transient decomposition module successfully isolates camera-attached contaminants into the transient field. This prevents the artifacts from leaking into the static field, allowing our system to output a clean, unobstructed reconstruction of the colon tissue while recovering the degraded ground truth via alpha blending.

\begin{figure}[!t]
\centering
\includegraphics[width=0.9\columnwidth]{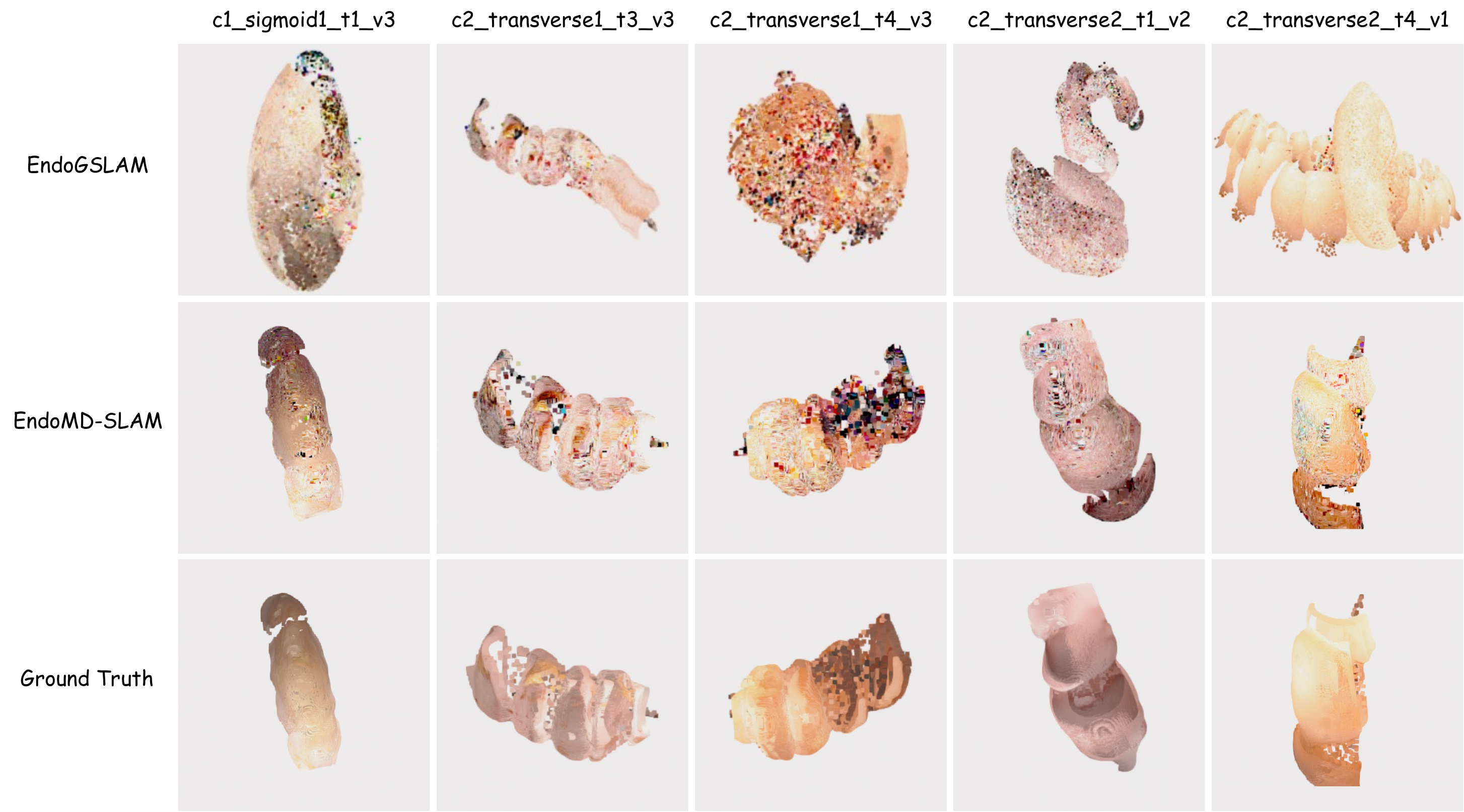}
\caption{\textbf{Qualitative 3D reconstruction of global colon anatomy under severe lens contamination.} The top row (EndoGSLAM) exhibits catastrophic structural breakdown and geometric distortion caused by the naive fusion of transient artifacts and tracking drift. Conversely, EndoMD-SLAM (middle row) successfully preserves the cohesive 3D anatomy, closely matching the Ground Truth (bottom row). This demonstrates the effectiveness of our memory and decomposition mechanisms in isolating contamination and gating erroneous map updates.}
\label{fig_qual_map}
\vspace{-3mm}
\end{figure}

Maintaining global map consistency under optical degradation requires explicit artifact handling, as visualized in Fig.~\ref{fig_qual_map}. Under severe optical degradation, EndoGSLAM exhibits catastrophic structural breakdown. Because the baseline lacks mechanisms to filter transient artifacts or pause updates during tracking failures, it forcefully integrates corrupted poses and 2D contaminants into the global map. This results in a massive accumulation of floating Gaussian artifacts and permanently warps the overall topology of the colon. Conversely, EndoMD-SLAM consistently recovers clean, coherent 3D structures. By leveraging the static-transient decomposition to protect the persistent static anatomy, our system accurately preserves the complex tubular structures and mucosal folds, closely matching the Ground Truth without the severe noisy accumulation observed in existing frameworks.

To evaluate tracking drift, Fig.~\ref{fig_qual_recovery} plots the estimated camera trajectories. Endoscopic procedures frequently involve sudden optical degradations, such as water flushing for lens cleaning, which severely break frame-to-frame visual continuity. Under these conditions, standard baselines suffer from frequent localization failures and their estimat ed trajectories wildly diverge from the Ground Truth. By leveraging our temporal memory module to safely gate map updates and perform failure-aware relocalization during these highly unreliable intervals, EndoMD-SLAM maintains robust pose estimation. Our system successfully recovers from complete visual loss, preventing long-term trajectory drift and ensuring the geometric integrity of the downstream mapping process.

\subsection{Runtime}
\label{sec_runtime}
We evaluate the computational efficiency of EndoMD-SLAM against competitive baselines on an NVIDIA RTX A6000 server, with quantitative results detailed in Tab.~\ref{tab_speed}. Our system achieves the highest overall throughput at 0.95 FPS, substantially outperforming the closest baseline, EndoGSLAM (0.65 FPS). Furthermore, our framework achieves both the fastest tracking at 0.170 seconds per frame and the fastest mapping at 0.874 seconds per frame. This superior efficiency is a direct result of our architecture specifically tailored for severe optical degradation. Our mapping process is highly efficient because the temporal memory and the decomposition of static and transient fields prevent degraded observations from polluting the persistent Gaussian map. By explicitly filtering out invalid points, our system keeps the representation extremely compact and drastically reduces the optimization cost. Concurrently, our tracking overhead remains exceptionally low due to our reliability gating mechanism. Instead of forcing the system to painstakingly iterate and track through regions with obviously low confidence, we immediately suspend standard tracking and switch directly to relocalization using our temporal memory. Bypassing these computationally heavy attempts to track unreliable frames ensures that both localization and mapping remain consistently fast throughout the entire procedure.

\begin{table}[t]
\centering
\small
\setlength{\tabcolsep}{4pt}
\renewcommand{\arraystretch}{1.12}
\begin{tabular}{l|c|cc}
\hline
\multirow{2}{*}{Method} & Throughput & Tracking Time & Mapping Time \\
 & (FPS) $\uparrow$ & (s/frame) $\downarrow$ & (s/frame) $\downarrow$ \\
\hline
MonoGS~\cite{matsuki2024gsslams}      & 0.25 & 1.807 & 2.139 \\
EndoGSLAM~\cite{wang2024endogslam}   & 0.65 & 0.574 & 0.973 \\
NICE-SLAM~\cite{zhu2022niceslam}     & 0.20 & 0.195 & 4.814 \\
\hline
\textbf{EndoMD-SLAM}                 & \textbf{0.95} & \textbf{0.170} & \textbf{0.874} \\
\hline
\end{tabular}

\caption{\textbf{Runtime efficiency. }Evaluated on an NVIDIA RTX A6000 server at identical input resolutions. By explicitly gating map updates and preventing the generation of redundant transient geometry, EndoMD-SLAM achieves the highest overall throughput and fastest mapping speed. Best results are highlighted in \textbf{bold}.}
\label{tab_speed}
\vspace{-5mm}
\end{table}

%% file: sec/5_conclusion.tex
\section{Conclusion}
\label{sec_conclusion}

In this work, we presented EndoMD-SLAM, an endoscopic Gaussian splatting SLAM framework designed to tackle the severe optical degradations routinely encountered in clinical procedures. We identified that standard systems struggle under optical degradation because they assume clean visual observations, leading to tracking drift and the erroneous fusion of 2D transient artifacts into the global 3D map. To overcome these coupled vulnerabilities, EndoMD-SLAM leverages complementary memory and decomposition mechanisms. Our static-transient decomposition isolates camera-attached contaminants into a dedicated transient field, preserving the structural integrity of the static anatomy. Concurrently, our temporal memory maintains a clean state history to safely gate map updates and enable drift-aware relocalization during highly degraded intervals. Extensive evaluations demonstrate that our approach significantly outperforms existing state-of-the-art baselines in both trajectory accuracy and rendering. While EndoMD-SLAM establishes a highly robust foundation for degradation-aware mapping, further elevating its rendering fidelity and optimizing computational efficiency for strictly real-time clinical deployment remain important avenues for future work.

%% file: refs.bib
@article{azagra2023endomapper,
  title={Endomapper dataset of complete calibrated endoscopy procedures},
  author={Azagra, Pablo and Sostres, Carlos and Ferr{\'a}ndez, {\'A}ngel and Riazuelo, Luis and Tomasini, Clara and Barbed, O Le{\'o}n and Morlana, Javier and Recasens, David and Batlle, Victor M and G{\'o}mez-Rodr{\'\i}guez, Juan J and others},
  journal={Scientific Data},
  volume={10},
  number={1},
  pages={671},
  year={2023},
  publisher={Nature Publishing Group UK London}
}

@inproceedings{ding2026extrinsplat,
  title={ExtrinSplat: Decoupling Geometry and Semantics for Open-Vocabulary Understanding in 3D Gaussian Splatting},
  author={Ding, Jiayu and Liu, Xinpeng and Pan, Zhiyi and Long, Shiqiang and Li, Ge},
  booktitle={Proceedings of the IEEE/CVF Conference on Computer Vision and Pattern Recognition},
  pages={31019--31028},
  year={2026}
}

@inproceedings{wang2024endogslam,
  title={Endogslam: Real-time dense reconstruction and tracking in endoscopic surgeries using gaussian splatting},
  author={Wang, Kailing and Yang, Chen and Wang, Yuehao and Li, Sikuang and Wang, Yan and Dou, Qi and Yang, Xiaokang and Shen, Wei},
  booktitle={International Conference on Medical Image Computing and Computer-Assisted Intervention},
  pages={219--229},
  year={2024},
  organization={Springer}
}

@inproceedings{wu2025endoflow,
  title={Endoflow-slam: Real-time endoscopic slam with flow-constrained gaussian splatting},
  author={Wu, Taoyu and Miao, Yiyi and Li, Zhuoxiao and Zhao, Haocheng and Dang, Kang and Su, Jionglong and Yu, Limin and Li, Haoang},
  booktitle={International Conference on Medical Image Computing and Computer-Assisted Intervention},
  pages={202--212},
  year={2025},
  organization={Springer}
}

@article{yoshida2017novel,
  title={A novel lens cleaner to prevent water drop adhesions during colonoscopy and esophagogastroduodenoscopy},
  author={Yoshida, Naohisa and Naito, Yuji and Yasuda, Ritsu and Murakami, Takaaki and Ogiso, Kiyoshi and Hirose, Ryohei and Inada, Yutaka and Dohi, Osamu and Okayama, Tetsuya and Kamada, Kazuhiro and others},
  journal={Endoscopy international open},
  volume={5},
  number={12},
  pages={E1235--E1241},
  year={2017},
  publisher={{\copyright} Georg Thieme Verlag KG}
}

@article{yoshida2015risk,
  title={Risk of lens cloudiness during colorectal endoscopic submucosal dissection and ability of a novel lens cleaner to maintain and restore endoscopic view},
  author={Yoshida, Naohisa and Naito, Yuji and Hirose, Ryohei and Ogiso, Kiyoshi and Siah, Kewin Tien Ho and Inada, Yutaka and Dohi, Osamu and Kamada, Kazuhiro and Katada, Kazuhiro and Uchiyama, Kazuhiko and others},
  journal={Digestive Endoscopy},
  volume={27},
  number={5},
  pages={609--617},
  year={2015},
  publisher={Wiley Online Library}
}

@article{golhar2025c3vdv2,
  title={C3VDv2--Colonoscopy 3D video dataset with enhanced realism},
  author={Golhar, Mayank V and Fretes, Lucas Sebastian Galeano and Ayers, Loren and Akshintala, Venkata S and Bobrow, Taylor L and Durr, Nicholas J},
  journal={arXiv preprint arXiv:2506.24074},
  year={2025}
}

@article{campos2021orb,
  title={Orb-slam3: An accurate open-source library for visual, visual--inertial, and multimap slam},
  author={Campos, Carlos and Elvira, Richard and Rodr{\'\i}guez, Juan J G{\'o}mez and Montiel, Jos{\'e} MM and Tard{\'o}s, Juan D},
  journal={IEEE transactions on robotics},
  volume={37},
  number={6},
  pages={1874--1890},
  year={2021},
  publisher={IEEE}
}

@article{richter2024realtimeendo3d,
  title={Advances in real-time 3D reconstruction for medical endoscopy},
  author={Richter, Alexander and Steinmann, Till and Rosenthal, Jean-Claude and Rupitsch, Stefan J},
  journal={Journal of imaging},
  volume={10},
  number={5},
  pages={120},
  year={2024},
  publisher={MDPI}
}

@article{elvira2024cudasiftslam,
  title   = {CudaSIFT-SLAM: Multiple-Map Visual SLAM for Full Procedure Mapping in Real Human Endoscopy},
  author  = {Elvira, Richard and Tard{\'o}s, Juan D. and Montiel, Jos{\'e} M. M.},
  journal = {arXiv preprint arXiv:2405.16932},
  year    = {2024}
}

@article{anadon2025densification,
  title={3D Densification for Multi-Map Monocular VSLAM in Endoscopy},
  author={Anad{\'o}n, X and Rodr{\'\i}guez-Puigvert, Javier and Montiel, JMM},
  journal={arXiv preprint arXiv:2503.14346},
  year={2025}
}

@inproceedings{huang2024endo4dgs,
  title={Endo-4dgs: Endoscopic monocular scene reconstruction with 4d gaussian splatting},
  author={Huang, Yiming and Cui, Beilei and Bai, Long and Guo, Ziqi and Xu, Mengya and Islam, Mobarakol and Ren, Hongliang},
  booktitle={International Conference on Medical Image Computing and Computer-Assisted Intervention},
  pages={197--207},
  year={2024},
  organization={Springer}
}

@article{recasens2021endodepthmotion,
  title={Endo-depth-and-motion: Reconstruction and tracking in endoscopic videos using depth networks and photometric constraints},
  author={Recasens, David and Lamarca, Jos{\'e} and F{\'a}cil, Jos{\'e} M and Montiel, JM Martinez and Civera, Javier},
  journal={IEEE Robotics and Automation Letters},
  volume={6},
  number={4},
  pages={7225--7232},
  year={2021},
  publisher={IEEE}
}

@inproceedings{batlle2023lightneus,
  title={Lightneus: Neural surface reconstruction in endoscopy using illumination decline},
  author={Batlle, V{\'\i}ctor M and Montiel, Jos{\'e} MM and Fua, Pascal and Tard{\'o}s, Juan D},
  booktitle={International Conference on Medical Image Computing and Computer-Assisted Intervention},
  pages={502--512},
  year={2023},
  organization={Springer}
}

@article{dunnbeltran2024nflba,
  title={NFL-BA: Near-Field Light Bundle Adjustment for SLAM in Dynamic Lighting},
  author={Beltran, Andrea Dunn and Rho, Daniel and Niethammer, Marc and Sengupta, Roni},
  journal={arXiv preprint arXiv:2412.13176},
  year={2024}
}

@article{engel2018dso,
  title={Direct sparse odometry},
  author={Engel, Jakob and Koltun, Vladlen and Cremers, Daniel},
  journal={IEEE transactions on pattern analysis and machine intelligence},
  volume={40},
  number={3},
  pages={611--625},
  year={2017},
  publisher={IEEE}
}

@inproceedings{zhu2022niceslam,
  title={Nice-slam: Neural implicit scalable encoding for slam},
  author={Zhu, Zihan and Peng, Songyou and Larsson, Viktor and Xu, Weiwei and Bao, Hujun and Cui, Zhaopeng and Oswald, Martin R and Pollefeys, Marc},
  booktitle={Proceedings of the IEEE/CVF conference on computer vision and pattern recognition},
  pages={12786--12796},
  year={2022}
}

@inproceedings{sandstrom2023pointslam,
  title={Point-slam: Dense neural point cloud-based slam},
  author={Sandstr{\"o}m, Erik and Li, Yue and Van Gool, Luc and Oswald, Martin R},
  booktitle={Proceedings of the IEEE/CVF international conference on computer vision},
  pages={18433--18444},
  year={2023}
}

@article{xu2024dgslam,
  title={Dg-slam: Robust dynamic gaussian splatting slam with hybrid pose optimization},
  author={Xu, Yueming and Jiang, Haochen and Xiao, Zhongyang and Feng, Jianfeng and Zhang, Li},
  journal={Advances in Neural Information Processing Systems},
  volume={37},
  pages={51577--51596},
  year={2024}
}

@article{kong2024dgsslam,
  title={Dgs-slam: Gaussian splatting slam in dynamic environment},
  author={Kong, Mangyu and Lee, Jaewon and Lee, Seongwon and Kim, Euntai},
  journal={arXiv preprint arXiv:2411.10722},
  year={2024}
}

@inproceedings{runz2017cofusion,
  title={Co-fusion: Real-time segmentation, tracking and fusion of multiple objects},
  author={R{\"u}nz, Martin and Agapito, Lourdes},
  booktitle={2017 IEEE International Conference on Robotics and Automation (ICRA)},
  pages={4471--4478},
  year={2017},
  organization={IEEE}
}

@inproceedings{guo2024freesurgs,
  title={Free-surgs: Sfm-free 3d gaussian splatting for surgical scene reconstruction},
  author={Guo, Jiaxin and Wang, Jiangliu and Kang, Di and Dong, Wenzhen and Wang, Wenting and Liu, Yun-hui},
  booktitle={International Conference on Medical Image Computing and Computer-Assisted Intervention},
  pages={350--360},
  year={2024},
  organization={Springer}
}

@inproceedings{zhu2024endogs,
  title={Endogs: Deformable endoscopic tissues reconstruction with gaussian splatting},
  author={Zhu, Lingting and Wang, Zhao and Cui, Jiahao and Jin, Zhenchao and Lin, Guying and Yu, Lequan},
  booktitle={International Conference on Medical Image Computing and Computer-Assisted Intervention},
  pages={135--145},
  year={2024},
  organization={Springer}
}

@inproceedings{matsuki2024gsslams,
  title={Gaussian splatting slam},
  author={Matsuki, Hidenobu and Murai, Riku and Kelly, Paul HJ and Davison, Andrew J},
  booktitle={Proceedings of the IEEE/CVF conference on computer vision and pattern recognition},
  pages={18039--18048},
  year={2024}
}

@article{grasa2013visual,
  title={Visual SLAM for handheld monocular endoscope},
  author={Grasa, Oscar G and Bernal, Ernesto and Casado, Santiago and Gil, Ismael and Montiel, JMM},
  journal={IEEE transactions on medical imaging},
  volume={33},
  number={1},
  pages={135--146},
  year={2013},
  publisher={IEEE}
}

@article{mahmoud2017slam,
  title={SLAM based quasi dense reconstruction for minimally invasive surgery scenes},
  author={Mahmoud, Nader and Hostettler, Alexandre and Collins, Toby and Soler, Luc and Doignon, Christophe and Montiel, Jose Maria Martinez},
  journal={arXiv preprint arXiv:1705.09107},
  year={2017}
}

@inproceedings{wang2019visual,
  title={Visual slam for bronchoscope tracking and bronchus reconstruction in bronchoscopic navigation},
  author={Wang, Cheng and Oda, Masahiro and Hayashi, Yuichiro and Kitasaka, Takayuki and Honma, Hirotoshi and Takabatake, Hirotsugu and Mori, Masaki and Natori, Hiroshi and Mori, Kensaku},
  booktitle={Medical Imaging 2019: Image-Guided Procedures, Robotic Interventions, and Modeling},
  volume={10951},
  pages={51--57},
  year={2019},
  organization={SPIE}
}

@article{mildenhall2022nerf,
  title={Nerf: Representing scenes as neural radiance fields for view synthesis},
  author={Mildenhall, Ben and Srinivasan, Pratul P and Tancik, Matthew and Barron, Jonathan T and Ramamoorthi, Ravi and Ng, Ren},
  journal={Communications of the ACM},
  volume={65},
  number={1},
  pages={99--106},
  year={2021},
  publisher={ACM New York, NY, USA}
}

@article{kerbl2023gaussiansplatting,
  title={3d gaussian splatting for real-time radiance field rendering.},
  author={Kerbl, Bernhard and Kopanas, Georgios and Leimk{\"u}hler, Thomas and Drettakis, George and others},
  journal={ACM Trans. Graph.},
  volume={42},
  number={4},
  pages={139--1},
  year={2023}
}

@inproceedings{wang2024desplat,
  title={DeSplat: Decomposed Gaussian splatting for distractor-free rendering},
  author={Wang, Yihao and Klasson, Marcus and Turkulainen, Matias and Wang, Shuzhe and Kannala, Juho and Solin, Arno},
  booktitle={Proceedings of the Computer Vision and Pattern Recognition Conference},
  pages={722--732},
  year={2025}
}

@inproceedings{lin2024hybridgs,
  title={HybridGS: Decoupling transients and statics with 2D and 3D gaussian splatting},
  author={Lin, Jingyu and Gu, Jiaqi and Fan, Lubin and Wu, Bojian and Lou, Yujing and Chen, Renjie and Liu, Ligang and Ye, Jieping},
  booktitle={Proceedings of the Computer Vision and Pattern Recognition Conference},
  pages={788--797},
  year={2025}
}

@inproceedings{leroy2024mast3r,
  title={Grounding image matching in 3d with mast3r},
  author={Leroy, Vincent and Cabon, Yohann and Revaud, J{\'e}r{\^o}me},
  booktitle={European conference on computer vision},
  pages={71--91},
  year={2024},
  organization={Springer}
}

@inproceedings{murai2025mast3rslam,
  title={Mast3r-slam: Real-time dense slam with 3d reconstruction priors},
  author={Murai, Riku and Dexheimer, Eric and Davison, Andrew J},
  booktitle={2025 IEEE/CVF Conference on Computer Vision and Pattern Recognition (CVPR)},
  pages={16695--16705},
  year={2025},
  organization={IEEE}
}

@article{wang2004image,
  title={Image quality assessment: from error visibility to structural similarity},
  author={Wang, Zhou and Bovik, Alan C and Sheikh, Hamid R and Simoncelli, Eero P},
  journal={IEEE transactions on image processing},
  volume={13},
  number={4},
  pages={600--612},
  year={2004},
  publisher={IEEE}
}

@inproceedings{zhang2018unreasonable,
  title={The unreasonable effectiveness of deep features as a perceptual metric},
  author={Zhang, Richard and Isola, Phillip and Efros, Alexei A and Shechtman, Eli and Wang, Oliver},
  booktitle={Proceedings of the IEEE conference on computer vision and pattern recognition},
  pages={586--595},
  year={2018}
}

@inproceedings{sturm2012benchmark,
  title={A benchmark for the evaluation of RGB-D SLAM systems},
  author={Sturm, J{\"u}rgen and Engelhard, Nikolas and Endres, Felix and Burgard, Wolfram and Cremers, Daniel},
  booktitle={2012 IEEE/RSJ international conference on intelligent robots and systems},
  pages={573--580},
  year={2012},
  organization={IEEE}
}

@article{ciuti2020frontiers,
  title={Frontiers of robotic colonoscopy: A comprehensive review of robotic colonoscopes and technologies},
  author={Ciuti, Gastone and Skonieczna-{\.Z}ydecka, Karolina and Marlicz, Wojciech and Iacovacci, Veronica and Liu, Hongbin and Stoyanov, Danail and Arezzo, Alberto and Chiurazzi, Marcello and Toth, Ervin and Thorlacius, Henrik and others},
  journal={Journal of clinical medicine},
  volume={9},
  number={6},
  pages={1648},
  year={2020},
  publisher={MDPI}
}

@inproceedings{liu2022sage,
  title={Sage: slam with appearance and geometry prior for endoscopy},
  author={Liu, Xingtong and Li, Zhaoshuo and Ishii, Masaru and Hager, Gregory D and Taylor, Russell H and Unberath, Mathias},
  booktitle={2022 International conference on robotics and automation (ICRA)},
  pages={5587--5593},
  year={2022},
  organization={IEEE}
}

@inproceedings{gomez2021sddefslam,
  title={SD-DefSLAM: Semi-direct monocular SLAM for deformable and intracorporeal scenes},
  author={G{\'o}mez-Rodr{\'\i}guez, Juan J and Lamarca, Jos{\'e} and Morlana, Javier and Tard{\'o}s, Juan D and Montiel, Jos{\'e} MM},
  booktitle={2021 IEEE international conference on robotics and automation (ICRA)},
  pages={5170--5177},
  year={2021},
  organization={IEEE}
}

@inproceedings{kaleta2024prendo,
  title={Pr-endo: Physically based relightable gaussian splatting for endoscopy},
  author={Kaleta, Joanna and Smolak-Dy{\.z}ewska, Weronika and Malarz, Dawid and Dall’Alba, Diego and Korzeniowski, Przemys{\l}aw and Spurek, Przemys{\l}aw},
  booktitle={International Conference on Medical Image Computing and Computer-Assisted Intervention},
  pages={391--401},
  year={2025},
  organization={Springer}
}

@article{teufel2024oneslam,
  title={OneSLAM to map them all: a generalized approach to SLAM for monocular endoscopic imaging based on tracking any point},
  author={Teufel, Timo and Shu, Hongchao and Soberanis-Mukul, Roger D and Mangulabnan, Jan Emily and Sahu, Manish and Vedula, S Swaroop and Ishii, Masaru and Hager, Gregory and Taylor, Russell H and Unberath, Mathias},
  journal={International Journal of Computer Assisted Radiology and Surgery},
  volume={19},
  number={7},
  pages={1259--1266},
  year={2024},
  publisher={Springer}
}

@inproceedings{cao2025coarsetofinegsslam,
  title={Coarse-to-fine real-time GS-SLAM from endoscopic RGBD data},
  author={Cao, Zhihao and Zhang, Yingkui and Qian, Yinling and Wang, Qiong},
  booktitle={Seventeenth International Conference on Digital Image Processing (ICDIP 2025)},
  volume={13709},
  pages={210--219},
  year={2025},
  organization={SPIE}
}

@inproceedings{wang2025vggt,
  title={Vggt: Visual geometry grounded transformer},
  author={Wang, Jianyuan and Chen, Minghao and Karaev, Nikita and Vedaldi, Andrea and Rupprecht, Christian and Novotny, David},
  booktitle={Proceedings of the Computer Vision and Pattern Recognition Conference},
  pages={5294--5306},
  year={2025}
}

@article{lin2025depth,
  title={Depth anything 3: Recovering the visual space from any views},
  author={Lin, Haotong and Chen, Sili and Liew, Junhao and Chen, Donny Y and Li, Zhenyu and Shi, Guang and Feng, Jiashi and Kang, Bingyi},
  journal={arXiv preprint arXiv:2511.10647},
  year={2025}
}

@article{maggio2025vggt,
  title={Vggt-slam: Dense rgb slam optimized on the sl (4) manifold},
  author={Maggio, Dominic and Lim, Hyungtae and Carlone, Luca},
  journal={arXiv preprint arXiv:2505.12549},
  year={2025}
}

@article{rodriguez2024nr,
  title={Nr-slam: Nonrigid monocular slam},
  author={Rodriguez, Juan J Gomez and Montiel, Jos{\'e} MM and Tardos, Juan D},
  journal={IEEE Transactions on Robotics},
  volume={40},
  pages={4252--4264},
  year={2024},
  publisher={IEEE}
}
